\documentclass[10pt,twocolumn,letterpaper]{article}

\usepackage{wacv}
\usepackage{times}
\usepackage{epsfig}
\usepackage{graphicx}
\usepackage{amsmath}
\usepackage{amssymb}

\usepackage{epsfig}
\usepackage{graphicx}
\usepackage{amssymb}
\usepackage{epsfig}
\usepackage{amsmath}
\usepackage{multirow}
\usepackage{booktabs}
\usepackage[ruled]{algorithm2e}
\usepackage{subcaption}

\SetKwInOut{Parameter}{parameter}
\SetKwData{Left}{left}
\SetKwData{This}{this}
\SetKwData{Up}{up}
\SetKwFunction{Union}{Union}
\SetKwFunction{FindCompress}{FindCompress}
\SetKwInOut{Input}{input}
\SetKwInOut{Output}{output}
\SetKwInput{KwInput}{Input}                
\SetKwInput{KwOutput}{Output}              

\def\wacvPaperID{0532} 

\wacvfinalcopy 

\ifwacvfinal
\def\assignedStartPage{1} 
\fi

\ifwacvfinal
\usepackage[breaklinks=true,bookmarks=false]{hyperref}
\else

\usepackage[pagebackref=true,breaklinks=true,colorlinks,bookmarks=false]{hyperref}
\fi

\ifwacvfinal
\else
\fi

\begin{document}

\title{ Mitigating Uncertainty of Classifier for Unsupervised Domain Adaptation}
\author{Shanu Kumar\\
Microsoft, India\\
{\tt\small shankum@microsoft.com}
\and
Vinod K Kurmi\\
IIT Kanpur\\
{\tt\small vinodkk@iitk.ac.in}
\and
Praphul Singh\\
Oracle, India\\
{\tt\small praphul.singh@oracle.com}
\and
Vinay P Namboodiri\\
University of Bath\\
{\tt\small vpn22@bath.ac.uk}
}

\maketitle

\begin{abstract}
Understanding unsupervised domain adaptation has been an important task that has been well explored. However, the wide variety of methods have not analyzed the role of a classifier's performance in detail. In this paper, we thoroughly examine the role of a classifier in terms of matching source and target distributions. We specifically investigate the classifier ability by matching a) the distribution of features, b) probabilistic uncertainty for samples and c) certainty activation mappings. Our analysis suggests that using these three distributions does result in a consistently improved performance on all the datasets. Our work thus extends present knowledge on the role of the various distributions obtained from the classifier towards solving unsupervised domain adaptation.

\end{abstract}
\vspace{-1em}
\section{Introduction}

Efficacy of using deep neural networks for solving a variety of problems in computer vision has been well established \cite{krizhevsky_NIPS2012,he2016deep,ren2015faster,chen2017deeplab}. These models are trained on large datasets and work in a variety of real-world settings. However, it has been shown \cite{tzeng_ICCV2015} that even after being trained on large datasets, these models suffer from inferior performance when they are used on data that differ for reasons such as background clutter, distribution of classes and illumination and pose conditions. To address this mismatch in performance, there has been a large body of work that aims to reduce the covariate shift by adapt features using unsupervised domain adaptation \cite{patel2015visual,wang2018deep}.
However, these techniques proposed so far have mainly ignored the response of the classifiers. They have focused on aligning the feature representations between the source and target domains that are input to the classifiers. In our work, we argue that such an approach under-utilizes the use of the classifier. Through this work, we analyse the role of the classifer and observe that carefully matching the classifier's response for source and target domains can result in consistent improvement in performance and lower uncertainty in the target domain. 


\begin{figure}[!]
\begin{subfigure}{0.23\textwidth}
  \centering
    \includegraphics[scale=0.2]{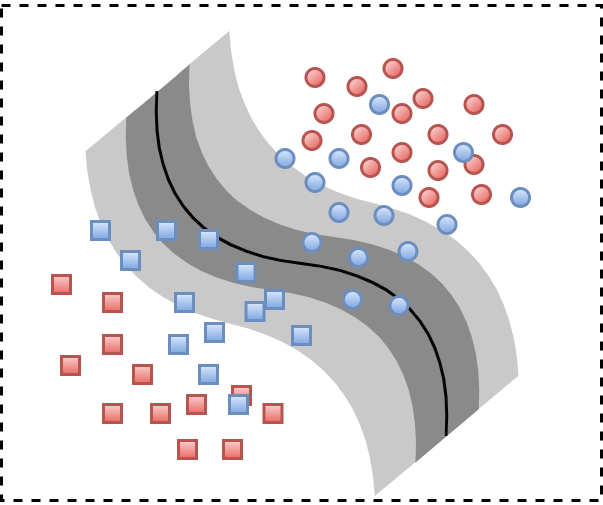}
            
  \caption{}
  \label{fig:intro_wrong}
\end{subfigure}
\begin{subfigure}{0.23\textwidth}
  \centering
    \includegraphics[scale=0.2]{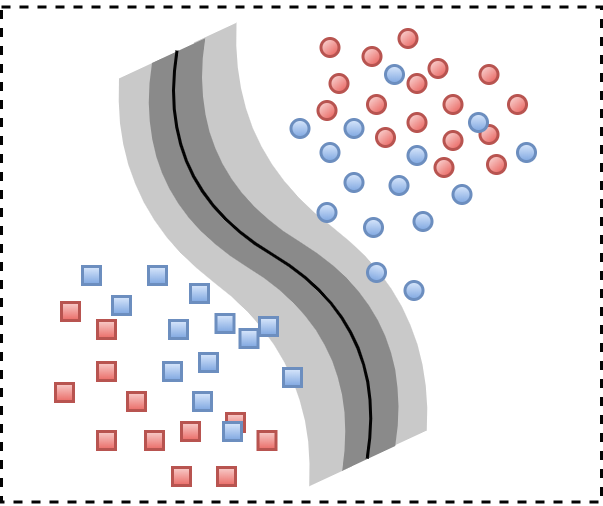}
  \caption{}
  \label{fig:intro_correct}
\end{subfigure}

\begin{subfigure}{0.5\textwidth}
  \centering
    \includegraphics[scale=0.24]{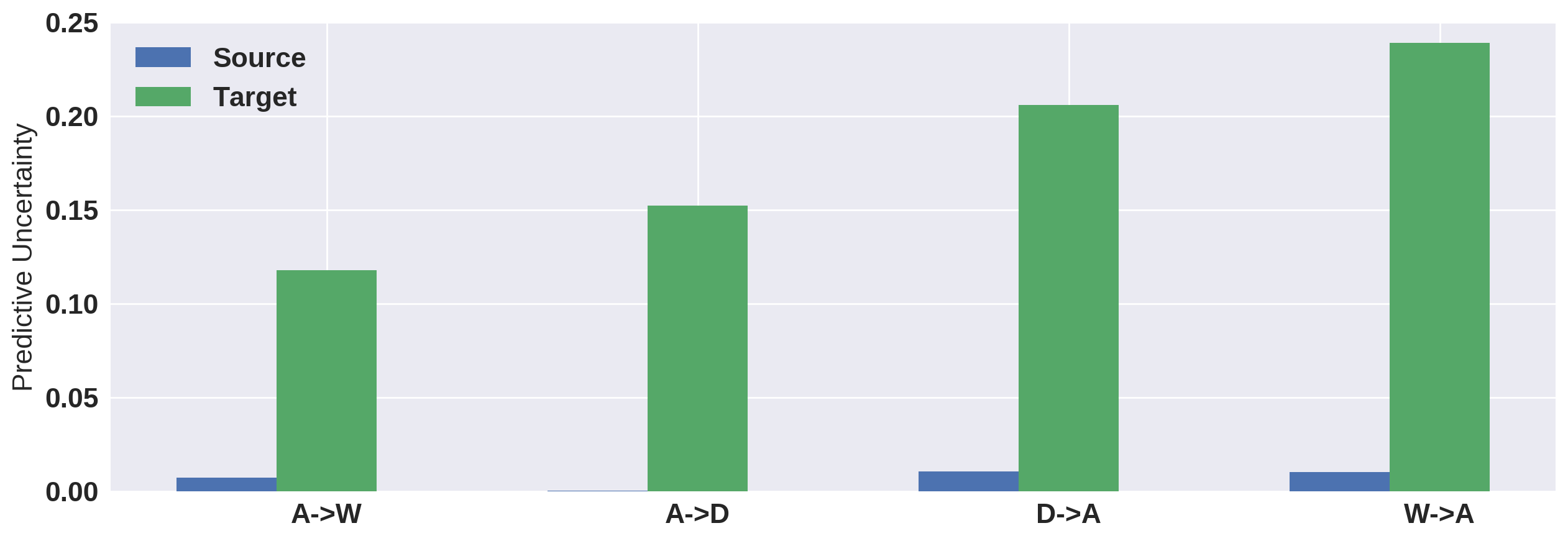}
  \caption{Predictive uncertainty}
  \label{fig:intro_uncer}
\end{subfigure}
\caption{Comparison of adversarial feature adaptation and our proposed method ({\color{red} red}: source domain and  {\color{blue} blue}: target domain). \textbf{(a)} The mean and variance of the decision boundary are not correctly approximated for the target domain, due to this all the samples of decision boundary do not separate most target domain examples. \textbf{(b)} Our approach learns better approximation of the mean and variance of the decision boundary and also learns more class-discriminative feature representations for the target domain. \textbf{(c)} Predictive uncertainty estimates of the target domain is relatively much higher than of the source domain images after only adapting the distribution of features on \textit{Office-31} dataset (AlexNet). For A$\rightarrow$D, uncertainty in the source domain is almost 0.}  
\label{fig:intro_fig}
\end{figure}

\begin{figure*}[!]
\begin{subfigure}{0.245\textwidth}
  \centering
  \captionsetup{justification=centering}
  \includegraphics[scale=0.14]{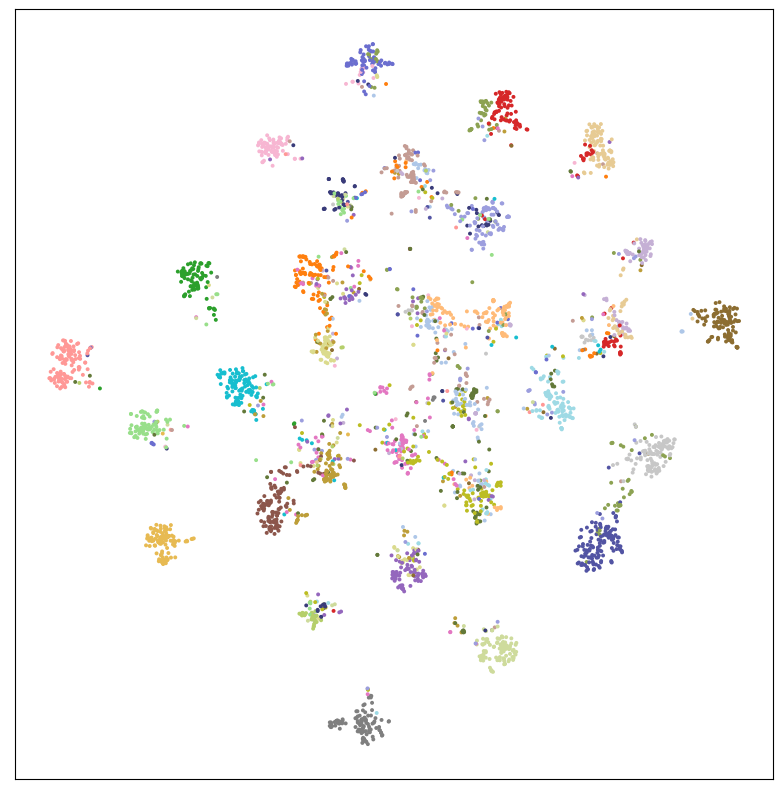}
  \caption{Class-wise Feature Representation}
  \label{fig:feat_cls_res}
\end{subfigure}
\begin{subfigure}{0.245\textwidth}
  \centering
  \captionsetup{justification=centering}
  \includegraphics[scale=0.14]{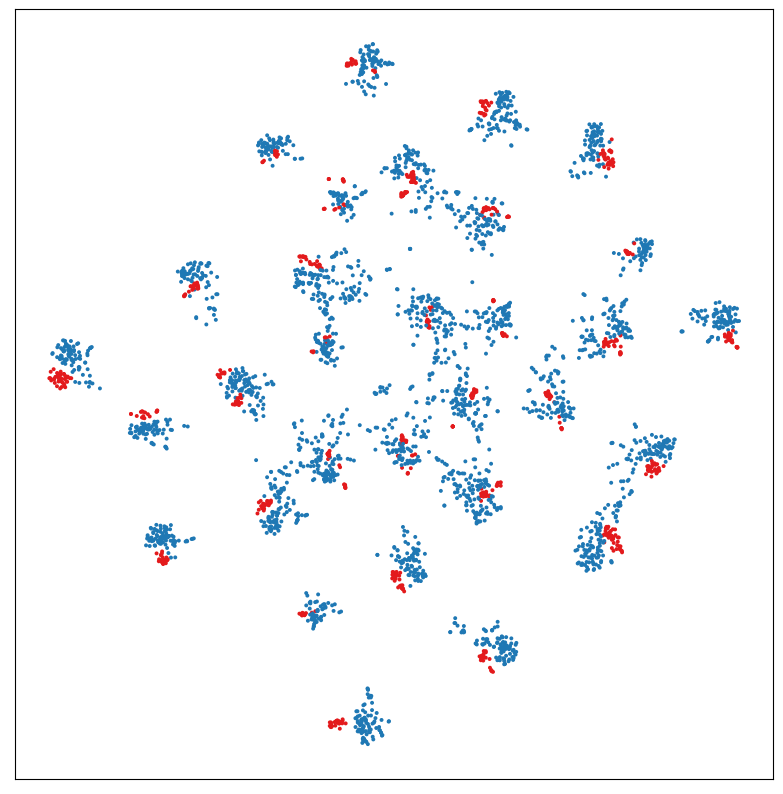}
  \caption{Domain-wise Feature Representation}
  \label{fig:feat_do_res}
\end{subfigure}
\begin{subfigure}{0.245\textwidth}
  \centering
  \captionsetup{justification=centering}
  \includegraphics[scale=0.14]{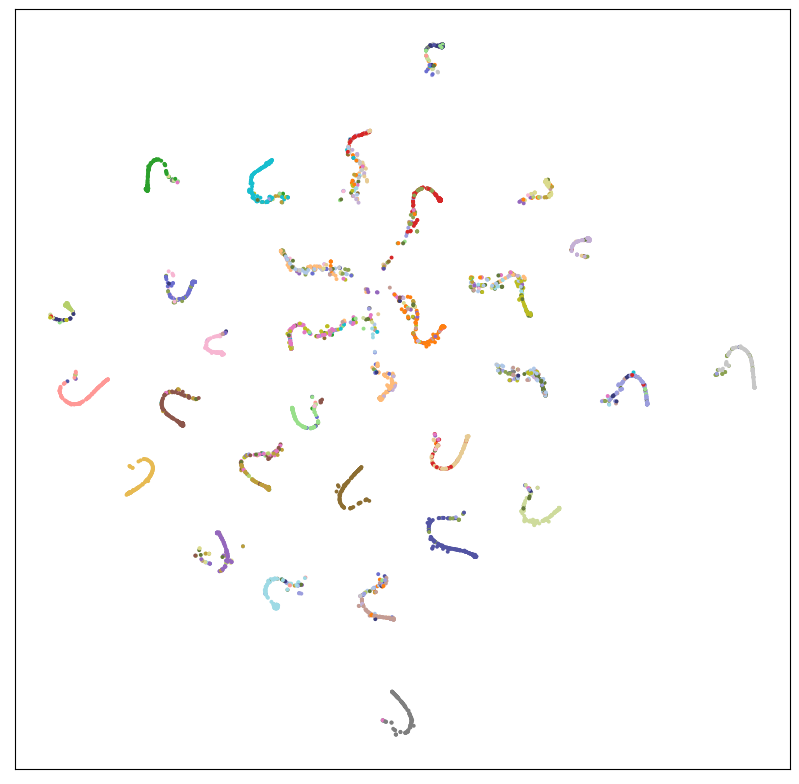}
  \caption{Class-wise Probability Distribution}
  \label{fig:prob_cls_res}
\end{subfigure}
\begin{subfigure}{0.245\textwidth}
  \centering
 \captionsetup{justification=centering}
  \includegraphics[scale=0.14]{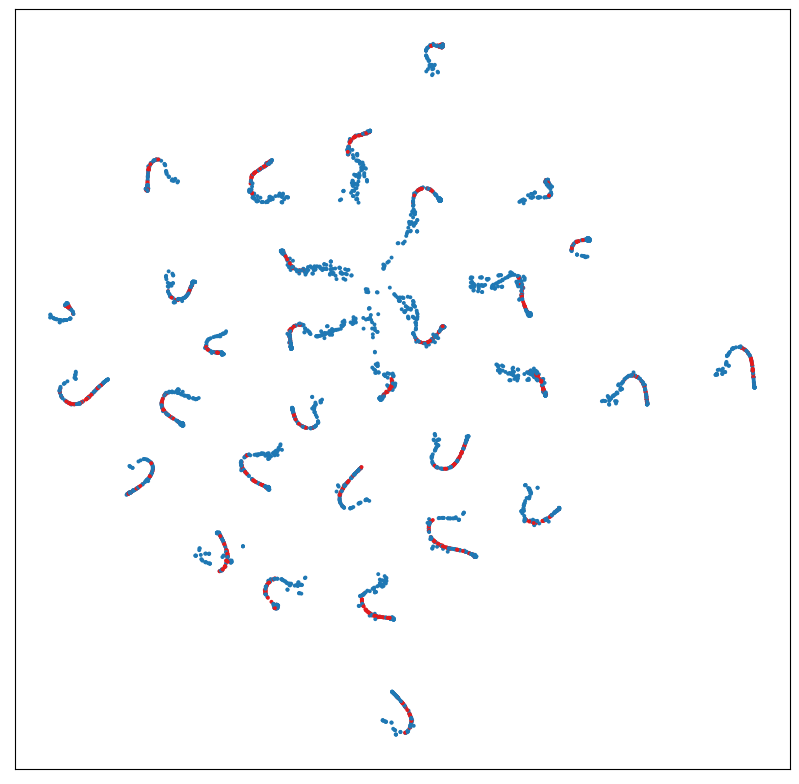}
  \caption{Domain-wise Probability Distribution}
  \label{fig:prob_do_res}
\end{subfigure}
\vspace{-0.5em}
\caption{Domain-wise and Class-wise t-SNE visualization of feature and probability distribution after preforming adversarial feature adaptation on Office31 dataset ({\color{red} Source}: W and {\color{blue} Target}: A) using AlexNet. The distribution of feature representation is domain-invariant after performing feature adaptation but many target domain features are not class-discriminative. In the probability distribution, we can observe domain shift between the target and source domain data points. We also have provided the magnified version of these plots in the supplementary material.}
\label{fig:prob_plot}
\vspace{-0.5em}
 
\end{figure*}

In order to understand the existing dominant approach adopted by most classifiers, we need to understand the underlying assumption. Most of the existing works in domain adaptation \cite{long2015learning,ganin2016domain,tzeng_CVPR2017} have focused on learning domain-invariant feature representations. They work on the assumption that the classifier learned in the source domain can be used in the target domain after feature adaptation. This theory is based on the principle that the posterior of the classifier in the target domain may be similar to the posterior learned in the source domain. But this may not be true always as the label-feature distribution can differ for the source and target domain. In figure \ref{fig:intro_wrong}, we illustrate an instance where the feature representations are domain-invariant and yet, the features in the target domain are not class-discriminative for the classifier learned in the source domain. As the classifier is only trained on the source domain, all the decision boundaries sampled from the posterior of the classifier entirely separates both classes in the source domain. Even though features are domain invariant, every sampled decision boundary do not separate both the classes in the target domain which results in lower performance and higher uncertainty. 
We observe similar pattern after performing adversarial adaptation of features for W$\rightarrow$A task using AlexNet. The features shown in the figure \ref{fig:feat_cls_res} are domain-invariant. In the figure \ref{fig:feat_do_res}, we can verify that features of the source domain images form unique class clusters but all the features of the target domain images are not class discriminative. The tools we use to analyze this performance are based on probabilistic uncertainty measures \cite{Gal2016Uncertainty,kendall2017uncertainties} and certainty activation mappings \cite{Kurmi_2019_CVPR}. Recently, there have been a few methods that have focused on classifier adaptation \cite{shu2018dirt, wen2019bayesian,liu2019transferable}.  \cite{shu2018dirt} addressed this issue by incorporating conditional entropy loss which refines the decision boundary, while \cite{liu2019transferable} enables the adaptation by generating transferable examples to fill in the gap between the source and target domains. However, we believe that they have not sufficiently addressed this problem.


In this paper, we solve the problem by learning a better approximation of the posterior of the classifier in the target domain. This is challenging as we cannot approximate the true posterior of the classifier without the class labels. We propose an unsupervised method that jointly aligns three specific distributions across domains. In our analysis, we observed that there is higher predictive uncertainty in the target domain as compared to the source domain. Also, there is a domain shift in the probability distribution of the classifier, which strengthens our belief that the mean of the posterior of the classifier is not correctly approximated in the target domain. Recently, \cite{Kurmi_2019_CVPR} has proposed certainty activation mappings that highlight the regions where the classifier is certain during predictions. We observe that certainty activation mappings in the source domain highlight class-discriminative regions, but in the target domain due to high uncertainty, these mappings do not highlight class-discriminative regions. This suggests that the approximation of the variance of the posterior is also not correct in the target domain. Based on these findings, we propose Triple Distribution Matching for Domain Adaptation (TDMDA), which jointly matches the distribution of feature representations, probability distributions, and certainty activation mappings from the target domain to that of the source domain. The main contributions of this paper are summarized as follows: 
\vspace{-0.5em}
\begin{itemize}
    \item We investigate the limitations of the traditional adversarial feature adaptation methods based on predictive uncertainty, probability distribution, and certainty activation mappings.
    \vspace{-0.5em}
    \item We propose TDMDA for learning a better approximation of the posterior of the classifier in the target domain by jointly matching three distributions.
    \vspace{-0.5em}
    \item We provide the evaluation of our method on various domain adaptation benchmark datasets and its comparisons with most of the state of the art methods.
    \vspace{-0.5em}
    \item We show the effectiveness of our method by providing an empirical analysis using the estimates of predictive uncertainty and various visualizations.
\end{itemize}

\section{Related Work}
\noindent\textbf{{Domain Adaptation:}}
There has been a wide variety of feature adaptation techniques for solving unsupervised domain adaptation. Early approaches were based on feature alignment through estimation of divergence between distributions such as  MMD~\cite{tzeng_arxiv2014,long2015learning,yan_CVPR2017}, CORAL~\cite{sun_ECCV2016}, CCD and Wasserstein metric~\cite{shen_AAAI2018wasserstein}. Ganin et al.\cite{ganin_ICML2015} enable the model to learn both class-discriminative and domain invariant representations by training a discriminator in an adversarial way. Other approaches such as \cite{shen_arxiv2017,tzeng_CVPR2017,tzeng_ICCV2015} have also proposed methods based on adversarial domain adaptation. More recent methods ~\cite{hoffman2018cycada, sankaranarayanan2018generate, Hu_2018_CVPR, bousmalis_2017CVPR} have performed unpaired image-to-image translation using generative adversarial networks (GAN) \cite{goodfellow2014generative} by generating synthetic images of one domain in other.
Attention based approaches have been also applied in domain adaptation.
TADA \cite{tada_aaai19} uses entropy of the local and global discriminators for finding transferable regions which are also used as attention for classifier.
In CADA~\cite{Kurmi_2019_CVPR}, attention is obtained by discriminator's certainty activation mappings and are used to focus on more discriminative regions. Huang \textit{et al}\cite{Huang_2018_ECCV} shows that there is domain discrepancy in the classifier by visualizing attention mechanism. They propose to align the distributions of activations of intermediate layers of source and target to adapt the classifier. We also propose a similar idea except the self attention or class activation mappings, we align certainty activation mappings.

Along with feature adaptation, few approaches have explored adaptation of classifier. Shu \textit{et al.}~\cite{shu2018dirt} propose decision-boundary refinement via conditional entropy loss. They have also incorporated domain adversarial training which penalizes violation of the cluster  assumption. \cite{liu2019transferable} encourages adaptation of classifier by generating transferable adversarial examples. Similarly, \cite{lee2019drop} uses adversarial dropout to learn discriminative features by pushing the decision boundary away from the target domain's features. \cite{deng2019cluster} also aligns the class conditional distribution of features of source and target domain. These have however, not studied classifier's distributions thoroughly as proposed in this work. We analyze it and provide a more general framework that considers various distributions for the classifier and 
jointly matches them to enforces both class-discriminative features and lower uncertainty in target domain.
\noindent\textbf{Uncertainty Estimation:} Our method relies on the various uncertainty estimation techniques that have been investigated. We provide an overview of the literature in this area. Bayesian neural networks (BNN) have been used in a wide variety of tasks for estimating predictive uncertainty. But determining the exact inference of $p(\mathbf{Y}|\mathbf{X})$ is an intractable problem because computing  the posterior expression $p(\theta|\mathbf{X}, \mathbf{Y}) = p(\mathbf{Y}|\mathbf{X}, \theta) p(\theta)/p(\mathbf{Y}|\mathbf{X})$ is analytically impossible. Thus the posterior $p(\theta|\mathbf{X}, \mathbf{Y})$ is approximated by a simple distribution $q^*(\theta)$ parameterized by $\{\mu, \sigma\}$. To approximate inference for deep learning models, \cite{gal2016dropout} has proposed a Dropout variational inference method. In \cite{ovadia2019can}, it has been shown that dropout variational inference has better calibration under in-domain distribution or distributional shifts than Temp Scaling \cite{guo2017calibration}, SVI\cite{blundell2015weight} and LL\cite{riquelme2018deep}. Thus, we have employed dropout variational inference for quantifying predictive uncertainty. 


Based on \cite{gal2016dropout}, Kurmi {\it et al.} \cite{Kurmi_2019_CVPR} has incorporated estimation of uncertainty for domain adaptation. They propose a method to visualize the activation mappings of certainty and uncertainty estimates. In their work, the discriminator's certainty activation mappings are used as an attention for guiding the classifier, but the classifier uncertainty estimates, and certainty activation mappings are not investigated. As a result, the method does not ensure the discriminativity of the adapted features. This drawback is thoroughly investigated and addressed in our work. ~\cite{wen2019bayesian} propose to minimize the difference of predictive uncertainty of the source and target domains, along with feature adaptation.

\section{Probabilistic Justification for the Proposed Approach}

In general, any  adversarial feature adaptation method focuses only on reducing the domain shift in the feature distribution, thus making $\mathcal{P}(F(\mathbf{X}_s)) \approx \mathcal{P}(F(\mathbf{X}_t))$. The inference in target domain can be defined mathematically as 
$\mathcal{P}(\mathbf{Y}_t | \mathbf{X}_t) = \mathcal{P}(\mathbf{Y}_t | F(\mathbf{X}_t)) \mathcal{P}(F(\mathbf{X}_t))$. Adapting the features results in $\mathcal{P}(F(\mathbf{X}_s)) \approx \mathcal{P}(F(\mathbf{X}_t))$. Now, $\mathcal{P}(\mathbf{Y}_t | F(\mathbf{X}_t))$ requires evaluation of the posterior $p(\theta_{c}|F(\mathbf{X}_t), \mathbf{Y}_t)$ for inference. Existing methods assume that the posterior of target-domain classifier can be approximated directly by using the posterior, $p(\theta_{c}|F(\mathbf{X}_s), \mathbf{Y}_s)$ learned from the source domain. But the approximation $q^*(\theta_c)$, learned by using the source domain do not guarantee that $p(\theta_{c}|F(\mathbf{X}_s), \mathbf{Y}_s)$ is the correct approximation of $p(\theta_{c}|F(\mathbf{X}_t), \mathbf{Y}_t)$ as the class labels for the target domain are not used. Therefore, reducing the domain shift alone do not guarantee that $\mathcal{P}(\mathbf{Y}_s | F(\mathbf{X}_s)) \approx \mathcal{P}(\mathbf{Y}_t | F(\mathbf{X}_t))$.


\subsection{High Predictive Uncertainty in Target Domain}
 In figure~\ref{fig:intro_uncer}, we have plotted the predictive uncertainty for source and target domains after the feature adaptation in Office-31 dataset~\cite{saenko_ECCV2010}. From the figure, it is evident that after the feature adaptation, there is reasonably high predictive uncertainty still present in the target domain. The reason for the high predictive uncertainty is that the parameters \{$\mu_c$, $\sigma_c$\} of the posterior $q^*(\theta_c)$, learned from the source domain are not the correct approximation of the posterior,  $p(\theta_{c}|F(\mathbf{X}_t), \mathbf{Y}_t)$ of the target domain. In the proposed method, we show that by considering the unlabeled target domain data $\mathbf{X}_t$, the posterior $q^*(\theta_c)$ can be approximated better for the target domain. The result for the same is provided later in figure~\ref{fig:cer_after}.
 

\subsection{Domain Shift in Probability Distribution}
Figure~\ref{fig:prob_plot} shows the probability distribution after performing the adversarial adaptation of features. In figures \ref{fig:prob_do_res} and \ref{fig:prob_cls_res}, we can see that all the data points lie on curved lines with unique class clusters in the source domain, but data points of the target domain do not follow a similar structure. Many data points in the target domain are in the center of the plot and do not correspond to any cluster. These are also not class-specific and causes a domain shift in the probability distribution $\hat{y}_c$ of the source and target domain. Domain shift can happen if either the mean of decision boundaries do not separate the data well in the target domain or the margin between the data and mean of decision boundaries is very small in the target domain. We can assume that this is a result of incorrect approximation of the $\mu_c$ of $q^*(\theta_c)$, as the classifier is trained only using source-domain labels. We can enforce better approximation of the mean, $\mu_c$ of $q^*(\theta_c)$ by minimizing the domain shift in the probability distribution $\hat{y}_c$, as we want the mean of decision boundaries to separate the classes in both domains very well with a large margin.

\begin{figure}[t]
    \centering
    \includegraphics[scale=0.23]{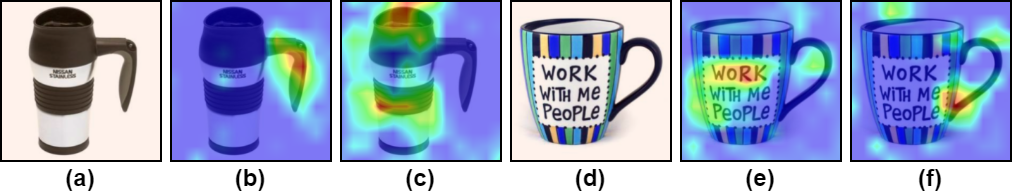}
    \caption{Visualization of certainty and uncertainty activation mappings of last convolutional layer in source domain. Uncertainty activation mappings are shown in (b) and (e) and, certainty activation mappings are shown in (c) and (f). Certainty activation mappings also highlight the class-discriminative regions of an image.}
  \label{fig:source_cer}
\end{figure}

\subsection{Certainty Activation Mapping}

\begin{figure*}[!]
    \centering
    \includegraphics[scale=0.23]{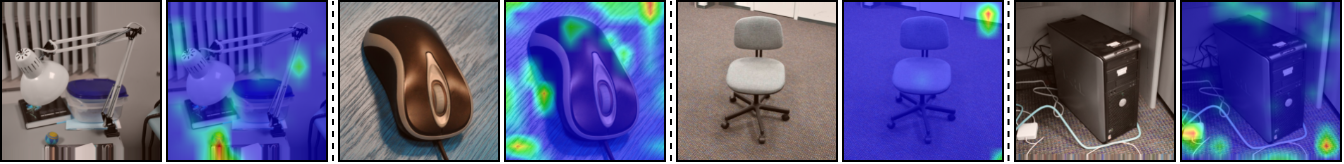}

    \caption{Visualization of certainty activation mappings of the last convolutional layer in target domain after adversarial feature adaptation for A$\rightarrow$D (AlexNet). These mappings are not highlighting any class-discriminative region.}
  \label{fig:dif_cer}
\end{figure*}

Similar to class activation mappings proposed in~\cite{selvaraju2017grad}, we can compute certainty activation mappings to identify the regions where the classifier is certain for performing predictions. Like class activation mappings, certainty activation mappings are highly class-discriminative for images of the source domain. Similarly, uncertainty activation mappings lead us to the uncertain regions for the predictions. By computing both the mappings through the approach proposed in~\cite{Kurmi_2019_CVPR},  we can visualize the certain and uncertain regions in the images for the classifier. This, in turn, provides much more information than the class activation mappings. In figure~\ref{fig:source_cer}, certainty and uncertainty activation mappings are shown respectively for the source domain for Office-31 dataset. In figure~\ref{fig:dif_cer}, certainty activation mapping in the target domain is shown after adversarial adaptation of features. We can observe that the class-discriminative regions are highlighted by uncertainty activation mappings instead of certainty activation mappings. This was not the case for the images in source domain as the posterior, $q^*(\theta_c)$ is approximated well for the source domain. For the target domain, we are using the same posterior, which therefore results in high uncertainty and non-class discriminative certainty activation mappings.


If the variance $\sigma_c$ of $q^*(\theta_c)$ is not learned correctly for the target domain, it will result in the incorrect sampling of decision boundaries. These decision boundaries may not separate the data well and thus will cause higher predictive uncertainty for target domain images. Variance for the target domain cannot be directly approximated, but reducing the uncertainty or any function proportional to uncertainty can lead to a better approximation of the variance $\sigma_c$. A similar kind of approach has been proposed in ~\cite{wen2019bayesian}, where the difference between the predictive uncertainties of source and target domain is minimized for producing consistent predictions. One important point to note is that reduction in predictive uncertainty do not necessarily mean a reduction in uncertainty on the class-discriminative regions. On the contrary, it may also be due to a higher reduction of uncertainty for the non-class discriminative regions. This brings us to an observation that learning correct explanation is better than learning correct answers.

We propose a mechanism to ameliorate the certainty activation mappings in the target domain to be class discriminative by matching the distribution of certainty activation mappings of the target domain to that of the source domain. As certainty activation mapping is a function of uncertainty, therefore this will lead to a reduction in uncertainty and finally to a better approximation of variance $\sigma_c$ of $q^*(\theta_c)$ in the target domain.

\begin{figure}[t]
     \centering
       \includegraphics[scale=0.1]{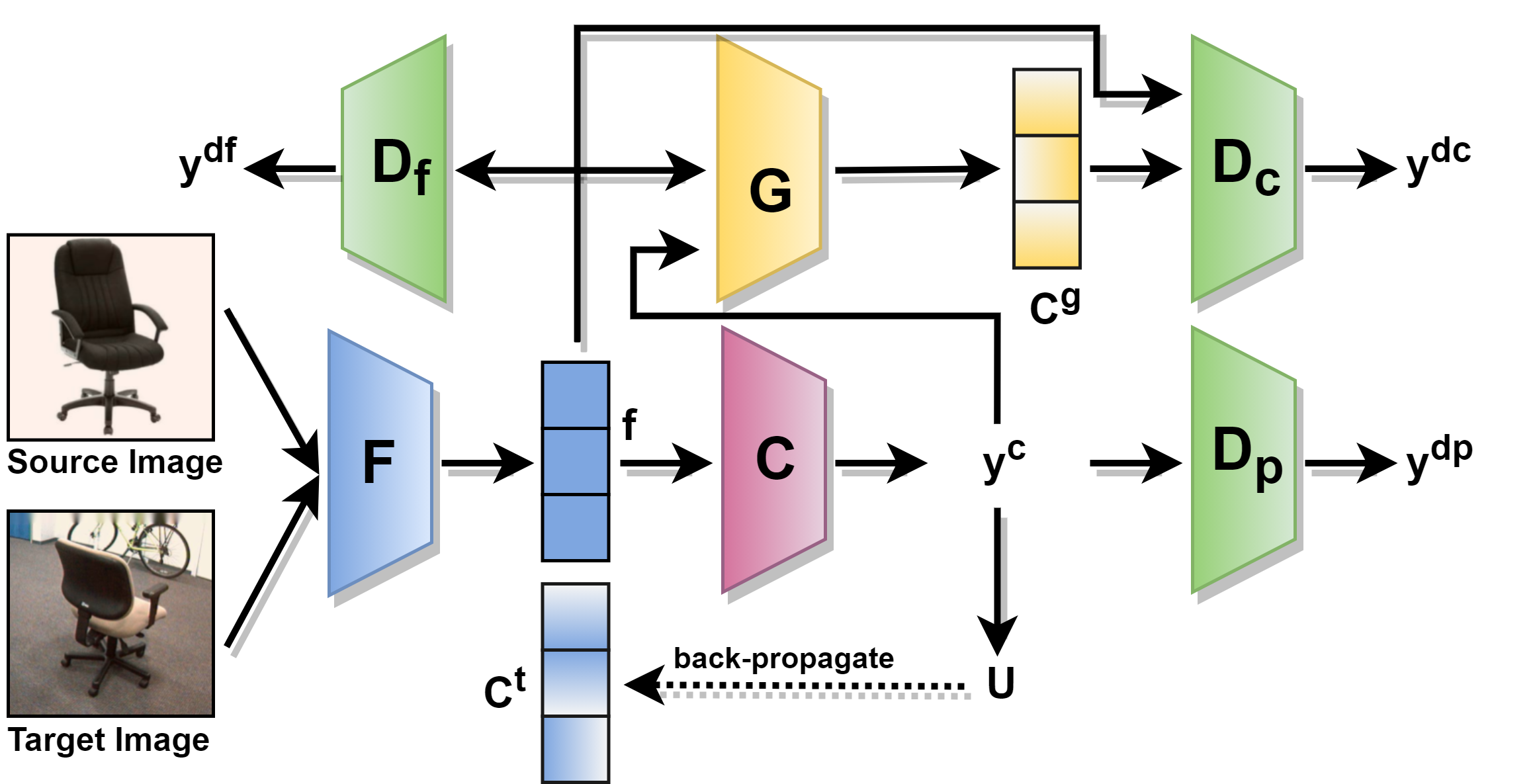}
  \caption{The architecture of Triple Distribution Matching for Domain Adaptation (TDMDA) consists of feature extractor $F$, Classifier $C$, Feature domain discriminator $D_f$, Probability distribution domain discriminator $D_p$, Certainty activation mapping domain discriminator $D_c$ and Certainty activation mapping generator $G$.}
      \label{fig:model}
 \end{figure}

\section{Methodology}
In unsupervised domain adaptation, we have a source domain $P(X_s, Y_s)$ with $n_s$ i.i.d. labeled observations ${\{x^s_i, y^s_i\}}^{n_s}_{i=1}$ and a target domain $P(X_t, Y_t)$ with $n_t$ i.i.d. unlabeled observations ${\{x^t_i\}}^{n_t}_{i=1}$. The goal is to learn transferable features $f = F(x)$ and an adaptive classifier $C$ which minimises the generalisation error on the target domain with lower uncertainty. For learning better approximation of the posterior $q^*(\theta_c)$ for target domain, we propose Triple Distribution Matching for Domain Adaptation (TDMDA), which solves the following three tasks: (a) Feature adaptation, (b) Predictive probability adaptation and (c) Certainty Activation Mapping adaptation.

\subsection{Feature Adaptation}

Adversarial adaptation of features aims to learn both class-discriminative and domain-invariant features. Domain-invariance is achieved by training a domain discriminator $D_f$, which predicts the domain label for both the source and target domain features, which are generated by training the feature extractor $F(x_i)$. The aim is to confuse the discriminator $D_f$ so that it is unable to distinguish between source and target domain, thus making the feature $f_i$ domain-invariant.
The classifier $C$ is trained to learn class discriminative features through the cross-entropy loss the source domain:
\begin{equation}
\label{eqn:loss_c}
\mathcal{L}_{c}= \frac{1}{n_s} \sum_{x_i \in \mathcal{D}_s} \mathcal{L}(C(\mathbf{f_i}), y_i^s)
\end{equation}
where $y_i^s$ is true class label. The domain discriminator $D_f$ is trained using the following loss:
\begin{equation}
\label{eqn:loss_df}
\mathcal{L}_{df}= \frac{1}{n_s+n_t} \sum_{x_i \in \mathcal{D}_s \cup \mathcal{D}_t} \mathcal{L}(D_f(\mathbf{f_i}), d_i)
\end{equation}
where $d_i$ is the domain label. The predictive uncertainty $U_{i}$ is estimated by using entropy of the average class probabilities of $T$ Monte Carlo samples of the features.
\begin{gather}
    p({y}_{i, t}^c) = \text{Softmax}(C(\mathbf{f_{i, t}})) \label{eqn:prob} \\ 
        U_{i} = -\frac{1}{\text{T}}  \sum_{t=1}^T \sum_{c=1}^C p({y}_{i, t}^c=c)*\log p({y}_{i, t}^c=c) \label{eqn:entropy} 
\end{gather}


\subsection{Probability Distribution Adaptation}

Earlier works used the mean of posterior $\mu_c$ learned from the source domain for the posterior of the target domain. For high-confidence predictions in the target domain, the $\mu_c$ for the target domain needs to be approximated separately using target domain images. Therefore, we aim at minimizing the domain shift in the probability distribution $C(\mathbf{f}_{i})$ of source and target domains for learning a better approximation of the mean $\mu_c$ of the posterior of the classifier. Similar to adversarial feature adaptation, we train a domain discriminator $D_{p}$ for predicting the domain of probability distribution $C(\mathbf{f}_{i})$. The classifier $C$, on the other hand, is trained to fool the discriminator by predicting high-confidence class predictions for both source and target domains. The probability distribution adaptation also results in learning more domain-invariant feature representation. The loss function for probability distribution adaptation is defined in Eq.~\ref{eqn:loss_dp}.
\begin{gather}
\mathcal{L}_{dp}= \frac{1}{n_s+n_t} \sum_{x_i \in \mathcal{D}_s \cup \mathcal{D}_t} \mathcal{L}(D_p(C(\mathbf{f_i})), d_i) \label{eqn:loss_dp}
\end{gather}

\subsection{Certainty Activation Mapping Adaptation}

We have followed ~\cite{Kurmi_2019_CVPR} for obtaining certainty activation mappings $\mathbf{C}_i^t$ for the classification task. We perform element-wise multiplication of feature with negative of the gradient of predictive uncertainty $U_i$ with respect to feature. We are only interested in the features that have a positive influence on the certainty estimates i.e., pixels whose intensity should be increased in order to improve the certainty. Therefore negative activations are replaced by a large negative number $b$, which will become zero after applying Softmax.
\begin{gather}
s_i= \begin{cases}
    \mathbf{f}_{i}* -\frac{\partial{U_{i}}}{\partial{{\mathbf{f}}_{i}}}  & \text{if } \mathbf{f}_{i}* -\frac{\partial{U_{i}}}{\partial{\mathbf{f}_{i}}} \ge 0\\
    b,              & \text{otherwise}
\end{cases}  \label{eqn:si} \\
 \mathbf{C}_i^t = 1 + \text{Softmax}(s_i) \label{eqn:cmap}
\end{gather}


Certainty activation mapping highlights the class-discriminative regions in the source domain. But, for many cases in the target domain, this is not observed, and this causes a domain shift for certainty activation mappings between source and target domains. Due to the wrong approximation of the variance $\sigma_c$ of the posterior of classifier $C$, there is high predictive uncertainty for the target domain as the classifier is uncertain on class-discriminative regions. To fix the issue, we propose to match the distribution of certainty activation mappings of the target domain to that of the source domain by using a domain discriminator $D_c$. The classifier is trained to fool the discriminator $D_c$ by learning a better approximation of the variance $\sigma_c$. 

The certainty activation mapping is a function of class logits $C(\mathbf{f}_{i})$ and features $\mathbf{f}_{i}$. To fool the discriminator $D_c$, we need to compute the gradients of predictive uncertainty with respect to certainty activation map, which itself is a function of derivative of predictive uncertainty. This will result in the computation of second derivatives of predictive uncertainty with respect to the features. We propose a way to deal with this complexity. As Neural Networks are universal approximators, we can easily learn the certainty activation mapping function using them. Therefore, we suggest using a generator $G$ for generating the certainty activation mappings for both source and target domains. Now, since we are generating the certainty activation mapping $\mathbf{C}_i^g$, we only have to deal with the first derivatives, thus making the computations simpler than before.
Certainty activation mapping generator $G$ is trained by minimizing the mean squared loss $\mathcal{L}_g$.
\begin{gather}
    \mathbf{C}_i^g=G(C(\mathbf{f}_{i}), \mathbf{f}_{i}) \\
    \mathcal{L}_{g}= ||\mathbf{C}_i^t-\mathbf{C}_i^g||^2
\label{eqn:loss_g} 
\end{gather}
Certainty activation mapping do not provide any information without the feature or image. Therefore the discriminator $D_c$ will have both certainty activation mappings $\mathbf{C}_i^g$ and $\mathbf{f}_{i}$ as inputs. Since we only aim to match the distribution of certainty activation mappings here, the features will act as weights for the certainty activation mappings. The discriminator $D_c$ is trained using the loss function $\mathcal{L}_{dc}$.
\begin{gather}
\mathcal{L}_{dc}= \frac{1}{n_s+n_t} \sum_{x_i \in \mathcal{D}_s \cup \mathcal{D}_t} \mathcal{L}(D_c(\mathbf{C}_i^g, \mathbf{f}_{i}), d_i) \label{eqn:loss_dc}
\end{gather}

\subsection{Final Objective}

We enable better approximation of the posterior $q^*(\theta_c)$ in both source and target domain by Triple Distribution Matching for Domain Adaptation (TDMDA), which jointly learns class discriminative and domain invariant features, domain invariant probability distribution, and class-discriminative certainty activation mappings. The final objective function $J$ of TDMDA is defined as:
\begin{multline}
\label{eqn:loss}
J(\theta_f, \theta_c, \theta_{df} ,{\theta}_g, \theta_{dp}, \theta_{dc}) = \mathcal{L}_c + \mathcal{L}_g -\lambda_{f} * \mathcal{L}_{df} \\
 -\lambda_{p} * \mathcal{L}_{dp}  -\lambda_{c} * \mathcal{L}_{dc}
\end{multline}

where $\theta_f, \theta_c, \theta_{df} ,{\theta}_g, \theta_{dp}, \theta_{dc}$ are the parameters of feature extractor $F$, classifier $C$, feature domain discriminator $D_f$, certainty activation mapping generator $G$, probability distribution domain discriminator $D_p$ and certainty activation mapping domain discriminator $D_c$ respectively. $\lambda_f, \lambda_p, \lambda_c$ are hyper-parameters that provides a trade-off between classifier and discriminators. The optimization problem is to find the parameters $\hat{\theta}_{f},\hat{\theta}_{c}, \hat{\theta}_{g}, \hat{\theta}_{df}, \hat{\theta}_{dp}, \hat{\theta}_{dc}$ that jointly satisfy:
\begin{equation}
\label{eqn:min}
    (\hat{\theta}_{f},\hat{\theta}_{c}, \hat{\theta}_{g}) =\underset{\theta_{f},\theta_{c}, \theta_{g}}{\text{arg min }} J(\theta_f, \theta_c, \theta_{df} ,{\theta}_g, \theta_{dp}, \theta_{dc})
\end{equation}

\begin{equation}
\label{eqn:max}
    (\hat{\theta}_{df}, \hat{\theta}_{dp}, \hat{\theta}_{dc}) =\underset{\theta_{df}, \theta_{dp}, \theta_{dc}}{\text{arg max }}J(\theta_f, \theta_c, \theta_{df} ,{\theta}_g, \theta_{dp}, \theta_{dc})
\end{equation}

  \begin{table}[!]
\centering
\caption {Classification accuracy (\%) on 
 \textit{Office-31} dataset for unsupervised domain adaptation (AlexNet~\cite{krizhevsky_NIPS2012})}
\vspace{-1em}
\begin{tabular}{cccccc}
\toprule
\textbf{Method }& A$\rightarrow$W & A$\rightarrow$D & D$\rightarrow$A & W$\rightarrow$A & Avg \\ 
  \midrule
 AlexNet\cite{krizhevsky_NIPS2012}  & 60.6  &64.2 & 45.5 & 48.3 & 54.65\\
  DANN\cite{ganin_ICML2015} & 73.0 & 72.3 & 52.4 & 50.4 & 62.03\\
  JAN\cite{long_ICML2017} & 75.2 & 72.8 & 57.5 & 56.3 & 65.45\\
 MADA\cite{pei_arxiv2018} & 78.5 &  74.1 & 56.0 & 54.5 & 65.78\\ 
  CDAN\cite{long_arxive2017conditional} & 77.9 & 74.6 & 55.1 & 57.5 & 66.28\\
   Entro\cite{wen2019bayesian} & 78.9 & 77.8 & 56.6 & 57.4 & 67.68 \\
     CAT\cite{deng2019cluster} & 80.7 & 76.4 & 63.7 & 62.2 & {\color{red}70.75} \\
   CADA\cite{Kurmi_2019_CVPR} & \textbf{83.4} & 80.1 & 59.8 & 59.5 & 70.70\\ 
 \midrule
TDMDA  & {82.6} & \textbf{85.2} &  \textbf{66.3} & \textbf{66.3} &  {\color{blue}\textbf{75.10}}\\
 \bottomrule 
\end{tabular}
  \label{tbl:office_alex}

 \end{table}
 
\begin{table}[!]
\centering
\caption {Classification accuracy (\%) on 
 \textit{Office-31} dataset for unsupervised domain adaptation (ResNet-50~\cite{he2016deep})}
\vspace{-1em}
\begin{tabular}{cccccc}
 \toprule
  \textbf{Method }& A$\rightarrow$W  & A$\rightarrow$D & D$\rightarrow$A & W$\rightarrow$A & Avg \\ 
  \midrule
 ResNet\cite{he2016deep} & 68.4 & 68.9 & 62.5 & 60.7 & 65.13\\
 DANN\cite{ganin_ICML2015} & 82.0 & 79.7 & 68.2 & 67.4 & 74.33 \\
 MADA\cite{pei_arxiv2018} & 90.0 & 87.8 & 70.3 & 66.4 & 78.62 \\
 DAAA~\cite{kang2018deep} & 86.8 & 88.8 & {74.3} & {73.9} & 80.95 \\
  CDAN\cite{long_arxive2017conditional} & 93.1 & 93.4 & 71.0 & 70.3 & 81.95  \\
 CAT\cite{deng2019cluster} & 94.4 & 90.8 & 72.2 & 70.2 & 81.90\\
  TAT\cite{liu2019transferable} & 92.5 & 93.2 & {73.1} & 72.1 & 82.73 \\
 TADA~\cite{tada_aaai19} &   94.2 & 92.8 & 72.6 & 73.6 & 83.30 \\
   CADA\cite{Kurmi_2019_CVPR} &  \textbf{97.0} & \textbf{95.6} & 71.5 & {73.1} & {\color{red}84.30} \\
      ETD\cite{li2020enhanced}  & 92.1 & 88.0 & 71.0 &  67.8 & 79.73\\
         DMRL\cite{wu2020dual}  & 90.8  & 93.4  &  73.0 & 71.2 & 82.10 \\
   ALDA\cite{chen2020adversarial}  &  95.6 & 94.0 &  72.2 & 72.5 & 83.58\\
   DADA\cite{tang2020discriminative}  &  92.3 & 93.9 &  74.4 & 74.2 & 83.78\\ 
 \midrule
 PMDA &  91.9 & 93.8 & 72.5 & 74.4 & 83.12\\
  CMDA &  93.6 & 94.1 & 73.3 & 74.5 & 83.88 \\
 TDMDA &   {94.8} & {93.5} & \textbf{74.7} & \textbf{75.8} & {\color{blue}\textbf{84.70}} \\
\bottomrule
\end{tabular}
  \label{tbl:office_res}
 \end{table}
 
  \begin{table}[!]
\centering
\caption {Classification accuracy (\%) on \textit{ImageCLEF} dataset for unsupervised domain adaptation (ResNet-50~\cite{he2016deep})} 
\setlength\tabcolsep{1pt}%
\vspace{-1em}
\begin{tabular}{cccccccc}
 \toprule
  \textbf{Method }& I $\rightarrow$ P & P$\rightarrow$ I &  I $\rightarrow$ C &C $\rightarrow$ I & C $\rightarrow$ P & P $\rightarrow$ C & Avg \\ 
  \midrule
ResNet\cite{he2016deep} & 74.8 & 83.9 & 91.5 & 78.0 & 65.5 & 91.2 & 80.7 \\
DANN\cite{ganin_ICML2015} & 75.0 & 86.0 & 96.2 & 87.0 & 74.3 & 91.5 & 85.0 \\
JAN\cite{long_ICML2017} & 76.8 & 88.0 &94.7 & 89.5 & 74.2 & 91.7 & 85.8 \\ 
MADA\cite{pei_arxiv2018} & 75.0 & 87.9 & 96.0 & 88.8 & 75.2 & 92.2 & 85.8 \\
CDAN\cite{long_arxive2017conditional} & 77.2 & 88.3 & \textbf{98.3} & 90.7 & 76.7 & 94.0 & 87.5  \\
CAT\cite{deng2019cluster} & 77.2 & 91.0 & 95.5 & 91.3 & 75.3 & 93.6 & 87.3\\
TAT\cite{liu2019transferable} & {78.8} & {92.0} & 97.5 & {92.0} & {78.2} & 94.7 & \color{red}88.9 \\
CADA\cite{Kurmi_2019_CVPR} & 78.0 & 90.5 & 96.7 & {92.0} &  77.2  &  95.5 & 88.3 \\

   DMRL\cite{wu2020dual} &77.3&  90.7 &  97.4&  91.8 &  76.0&  94.8&  88.0 \\
   
  AADA\cite{yangmind} &  \textbf{79.2} &\textbf{92.5} & 96.2& 91.4 &76.1 &94.7 &88.4 \\

 \midrule
TDMDA & 77.8 & {92.0} & 97.2 & \textbf{92.7} &  \textbf{78.7}  &  \textbf{96.2} & {\color{blue}{\textbf{89.1}}} \\
\bottomrule
\end{tabular}
  \label{tbl:imageclef}

 \end{table}

 \begin{table*}[ht]
 \centering
\caption {Classification accuracy (\%) on \textit{Office Home} dataset for unsupervised domain adaptation (ResNet-50~\cite{he2016deep})} 
\setlength\tabcolsep{0.7pt}%
\vspace{-1em}
\begin{tabular}{*{200}{c}}
 \toprule
  \textbf{Method}& Ar$\rightarrow$Cl & Ar$\rightarrow$Pr &  Ar$\rightarrow$Rw & Cl$\rightarrow$Ar & Cl$\rightarrow$Pr & Cl$\rightarrow$Rw & Pr$\rightarrow$Ar & Pr$\rightarrow$Cl & Pr$\rightarrow$Rw & Rw$\rightarrow$Ar & Rw$\rightarrow$Cl & Rw$\rightarrow$Pr & Avg \\ 
  \midrule
 ResNet\cite{he2016deep} & 34.9 & 50.0& 58.0 & 37.4 & 41.9 & 46.2 & 38.5 &31.2 & 60.4 & 53.9 & 41.2 & 59.9 & 46.1\\
DANN\cite{ganin_ICML2015} &45.6&59.3 & 70.1&47.0 &58.5 &60.9 &46.1 &43.7 &68.5 &63.2 &51.8 &76.8 &57.6 \\
JAN\cite{long_ICML2017} & 45.9 & 61.2 &68.9 &50.4 & 59.7 & 61.0 &45.8 &43.4&70.3 &63.9 &52.4 &76.8 &58.3 \\
CDAN\cite{long_arxive2017conditional} & 50.6 & 65.9 & 73.4 & 55.7 & 62.7 & 64.2 & 51.8 & 49.1 & 74.5 & 68.2 & 56.9 & 80.7 & 62.8\\
TAT\cite{liu2019transferable} & 51.6 & 69.5 & 75.4 & 59.4 & 69.5 & 68.6 & 59.5 & 50.5 & 76.8 & 70.9 & 56.6 & 81.6 & 65.8 \\
TADA\cite{tada_aaai19} & 53.1 & 72.3 & 77.2 & 59.0 & 71.2 & 72.1 & 59.7 & {53.1} & 78.4 & 72.4 & {60.0} & 82.9 & 67.6 \\
CADA\cite{Kurmi_2019_CVPR} & \textbf{56.9} & {76.4} & {80.7} & {61.3} & {75.2} & {75.2} & 63.2 & \textbf{54.5} & {80.7} & {73.9} & \textbf{61.5} & 84.1 & {70.2}\\

 ALDA\cite{chen2020adversarial} &  53.7 & 70.1 & 76.4& 60.2& 72.6& 71.5& 56.8& 51.9& 77.1& 70.2 &56.3& 82.1& 66.6 \\
  AADA\cite{yangmind} & 54.0  &  71.3  & 77.5  & 60.8  & 70.8  & 71.2  & 59.1   &51.8   & 76.9  & 71.0  & 57.4  & 81.8   &  67.0\\
   ETD\cite{li2020enhanced} & 51.3 & 71.9  & \textbf{85.7} & 57.6  & 69.2  & 73.7  & 57.8   &51.2  & 79.3 & 70.2 & 57.5 & 82.1  & 67.3  \\
  DCAN\cite{li2020domain} & 54.5  & 75.7   & 81.2   & \textbf{67.4}   & 74.0 & 76.3  & 67.4  & 52.7 &  80.6  & 74.1 &  59.1  & 83.5  &  {\color{red}70.5} \\
  
\midrule 
TDMDA & {55.4} & \textbf{79.3} & {81.6} &  {64.3} & \textbf{76.5} & \textbf{76.8} & \textbf{67.7} & {52.2} & \textbf{83.5} & \textbf{74.9} & {60.4} & \textbf{85.6} & \color{blue}\textbf{71.5}\\
 \bottomrule
\end{tabular}
  \label{tbl:home_office_res}
\vspace{-0.5em}
 \end{table*}
 
\section{Experiments and Results}

\subsection{Datasets}

\textbf{Office-31}~\cite{saenko_ECCV2010} dataset consists of images of office environment of three domains (Amazon, Webcam, and DSLR). It consists of 4,652 images and 31 classes. We have evaluated our model on four adaptation tasks (A$\rightarrow$W, A$\rightarrow$D, W$\rightarrow$A, D$\rightarrow$A). We removed the other two adaptation tasks (D$\rightarrow$W, W$\rightarrow$D) because they are very similar domains. 

\textbf{Office-Home}~\cite{venkateswara_cvpr2017deep} dataset consists of 65 classes with four domains Artistic (Ar), Clipart (Cl), Product (Pr) and Real-World (Rw) images. These domains are very different from each other, and adaptation is very challenging. We have reported all 12 transfer tasks for this dataset.

\textbf{ImageCLEF-2014} dataset consists of three datasets: Caltech-256 (C), ILSVRC 2012 (I), and Pascal VOC 2012 (P). There are 12 common classes, and each class has 50 samples. So, there is a total of 600 images in each domain.
 
\subsection{Results}

We have reported the results of the Office-31 dataset based on AlexNet in Table~\ref{tbl:office_alex}. Our method TDMDA has outperformed all of the mentioned methods by a significant margin of 4.35$\%$. Entro~\cite{wen2019bayesian} proposed to match the predictive uncertainty for adapting the classifier in the target domain. However, our method exceeds Entro by a large margin of 7.42$\%$ on the average accuracy. We have also reported our model performance for Office-31 dataset based on ResNet-50 in  Table \ref{tbl:office_res} with an improvement of 0.3$\%$. TDMDA outperforms methods based on classifier adaptation such as TAT\cite{liu2019transferable}, CAT\cite{deng2019cluster} and DAAA\cite{kang2018deep} by margin greater than 2$\%$. The proposed method even performs better than the recent approaches \cite{li2020domain, yangmind, li2020enhanced, chen2020adversarial} and it also incorporates complex attention mechanisms such as \cite{Kurmi_2019_CVPR, tada_aaai19}. This shows that the adaptation of classifiers in the target domain is necessary for effective domain adaptation. 

Table \ref{tbl:home_office_res} shows the results on Office-Home dataset. This dataset is far more challenging than other datasets as domain shift is significantly high here. Despite the challenge, our method achieves a substantial improvement over the other methods across most of the tasks. TDMDA outperforms TAT\cite{liu2019transferable} by a significant margin of 5.7$\%$ on average accuracy. While CADA\cite{Kurmi_2019_CVPR} outperforms our method when Clipart (Cl) is the target domain, but the results are still competitive performing on average better than CADA. 
Moreover, the proposed work can be plugged with any domain adaptation framework to improve the performance further. We have also reported the results on the ImageCLEF dataset in Table \ref{tbl:imageclef}. Our method exceeds the rest of the methods' performance with a minor improvement of 0.2$\%$ on average accuracy. The relatively small margin in improvement is because of the smaller domain shift in the dataset.  

\subsection{Ablation Study}

\begin{figure}[!]
     \centering
       \includegraphics[scale=0.25]{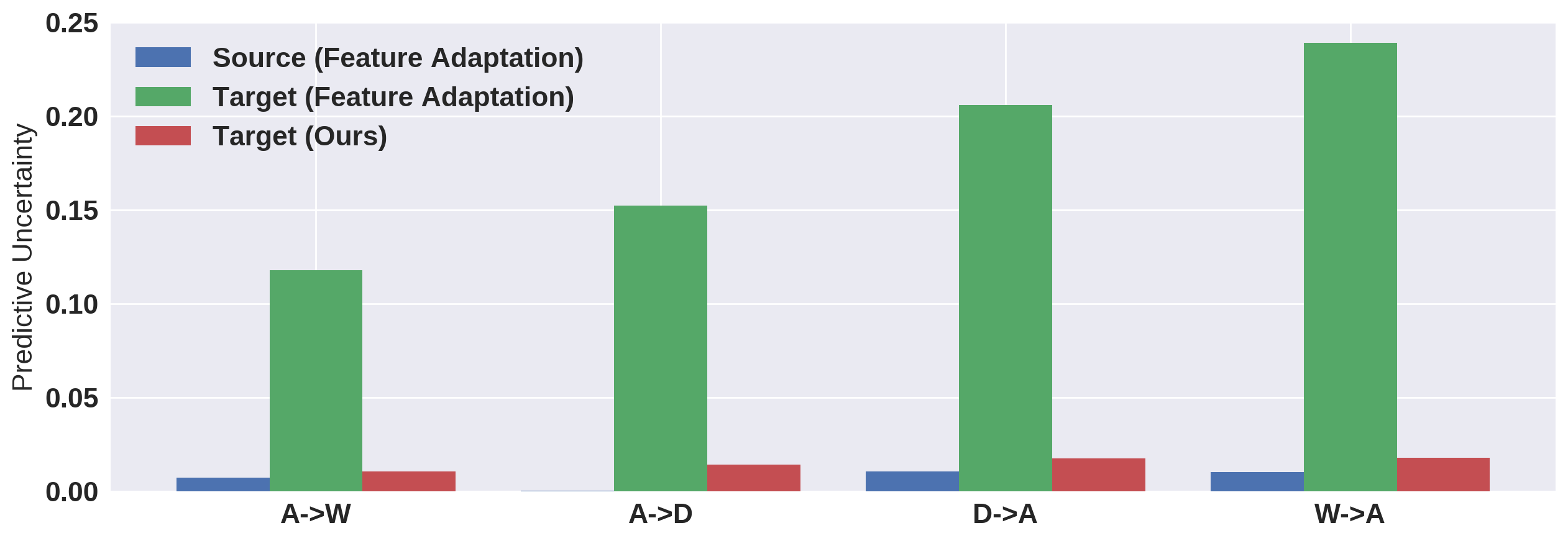}
       \caption{Predictive uncertainty in the target domain is significantly lower using our method than the adversarial feature adaptation method on \textit{Office-31} dataset (AlexNet)}
      \label{fig:cer_after}
  \vspace{-1em}
 \end{figure}
 
 \begin{figure*}[!]
\begin{subfigure}{0.24\textwidth}
  \centering
   \captionsetup{justification=centering}
  \includegraphics[scale=0.15]{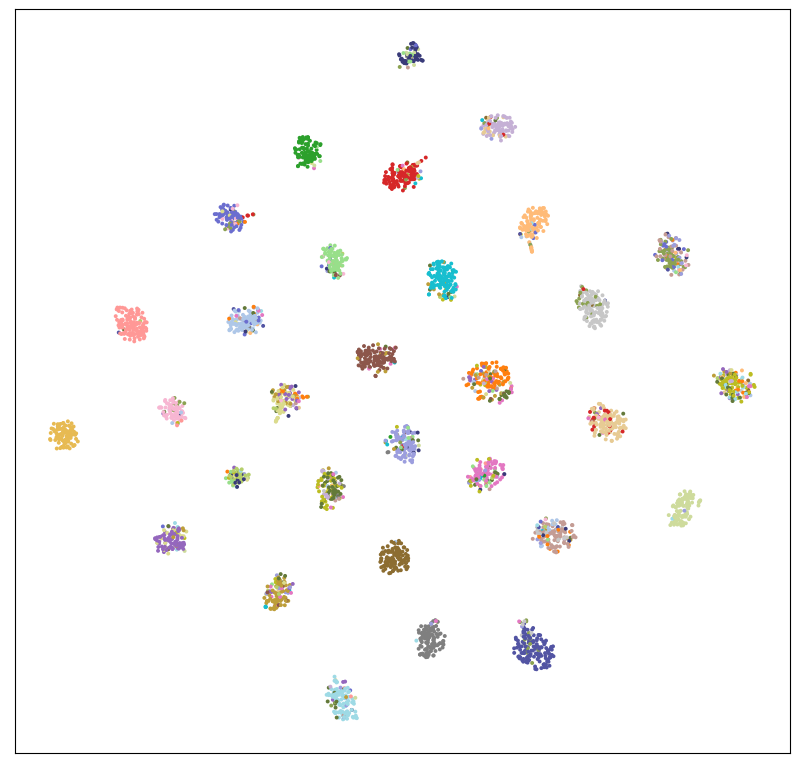}
  \caption{Class-wise Feature Representation}
  \label{fig:feat_after_cl}
\end{subfigure}
\begin{subfigure}{0.24\textwidth}
  \centering
   \captionsetup{justification=centering}
  \includegraphics[scale=0.15]{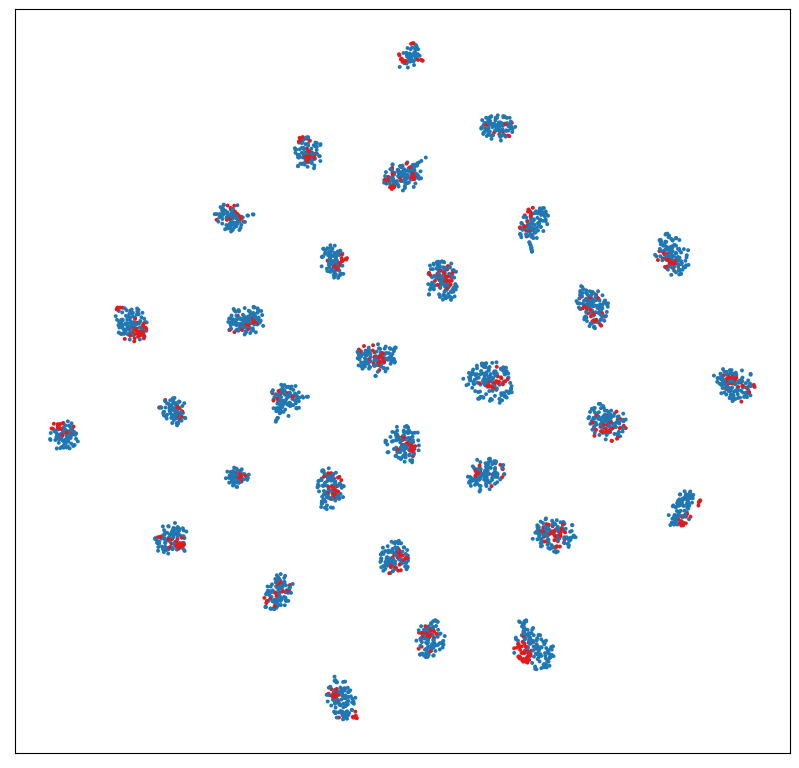}
  \caption{Domain-wise Feature Representation}
  \label{fig:feat_after_dp}
\end{subfigure}
\begin{subfigure}{0.24\textwidth}
  \centering
   \captionsetup{justification=centering}
  \includegraphics[scale=0.15]{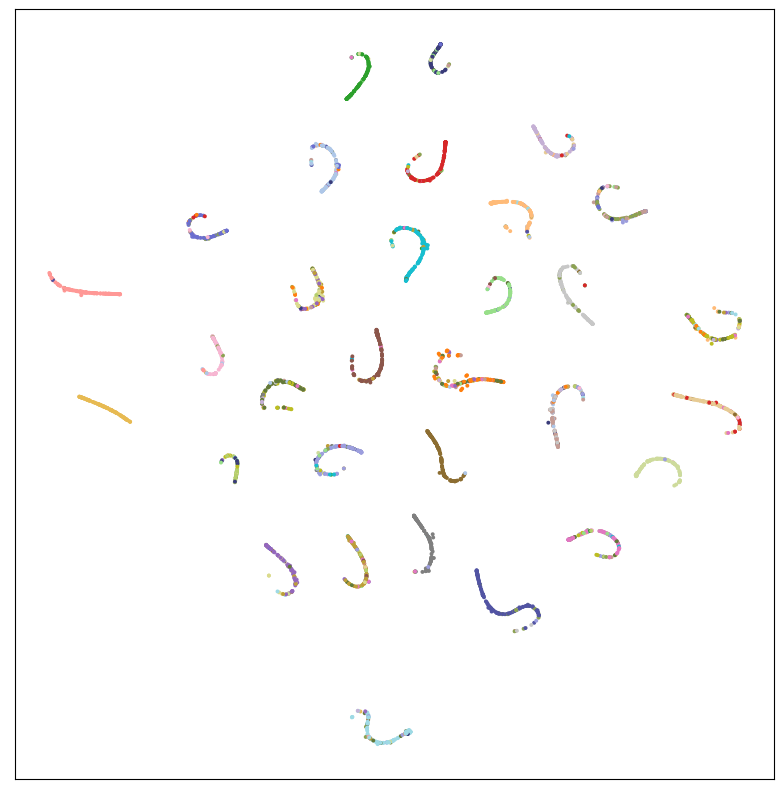}
  \caption{Class-wise Probability Distribution}
  \label{fig:prob_after_cl}
\end{subfigure}
\begin{subfigure}{0.24\textwidth}
  \centering
   \captionsetup{justification=centering}
  \includegraphics[scale=0.15]{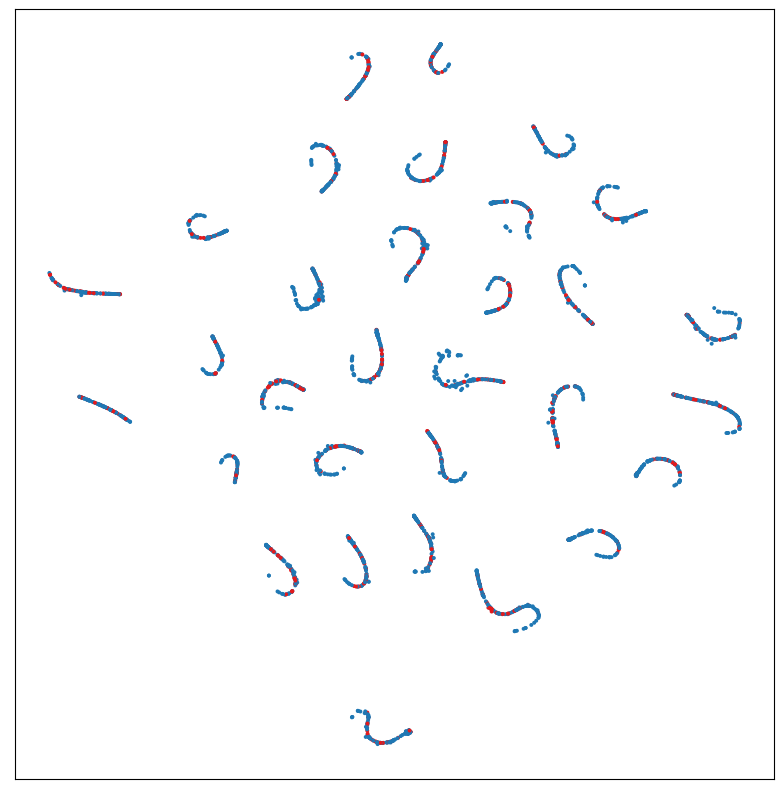}
  \caption{Domain-wise Probability Distribution}
  \label{fig:prob_after_dp}
\end{subfigure}
 \vspace{-0.5em}
\caption{(a) and (b) shows t-SNE visualization of feature representations learned by TDMDA which are now class-discriminative in the target domain. (c) and (d) represents t-SNE visualization of probability distribution of TDMDA without any domain shift for Office31 dataset ({\color{red} Source}: W and {\color{blue} Target}: A) using AlexNet.} 
\label{fig:feat_plot}
  \vspace{-0.5em}
\end{figure*}
We have also studied the contributions of both probability distribution adaptation and certainty activation mappings adaptation. The comparison between TDMDA and its variants is reported in Table \ref{tbl:office_res} for the Office-31 dataset. PMDA refers to the probability distribution adaptation, and CMDA refers to the certainty activation mapping adaptation. Results show that both PMDA and CMDA outperform most of the comparison methods. PMDA performs slightly worse than CMDA as it only encourages the classifier to learn class-discriminative feature representation with higher margins. This can lead to features aligned to incorrect class-cluster. By adapting both probability distribution along with certainty activation mapping reduces this problem. Thus, TDMDA exceeds both of them, which shows that learning both mean and variance is essential for approximating the classifier's posterior in the target domain.  

\section{Empirical Analysis}
\subsection{Predictive Uncertainty}
In figure~\ref{fig:cer_after}, we have shown the predictive uncertainty of source and target domain after feature adaptation, and target domain after using our method (TDMDA) on the Office-31 dataset using AlexNet. We can observe that predictive uncertainty on the target domain using our approach is much less than on target domain after feature adaptation for all the tasks. This shows that the posterior of the classifier is now much better approximated for the target domain. 

\subsection{Feature Visualization}
We have visualized the feature representations learned by our method TDMDA and performing only adversarial feature adaptation using t-SNE embedding~\cite{maaten2008visualizing} in figure \ref{fig:feat_after_dp} and \ref{fig:feat_do_res} for W$\rightarrow$A task using AlexNet respectively. Our method learns more class-discriminative feature representation and aligns most of the target domain images to a class cluster compared with features learned by feature adaptation, as observed in the figure \ref{fig:feat_after_cl}.
  
\subsection{Probability Distribution Visualization}
We have visualized source and target probability distributions on W$\rightarrow$A task on AlexNet using t-SNE embedding~\cite{maaten2008visualizing} in figures \ref{fig:prob_after_cl} and \ref{fig:prob_after_dp}. Most of the target domain's probability distributions are now aligned to a class-cluster similar to source domain. Now there is much lesser domain shift as compared to that shown in figure~\ref{fig:prob_plot}. This provides an evidence that our method has learned a better approximation of mean $\mu_c$ of the posterior for target domain.

\subsection{Certainty Activation Mapping}
\begin{figure}[t]
    \centering
    \includegraphics[scale=0.13]{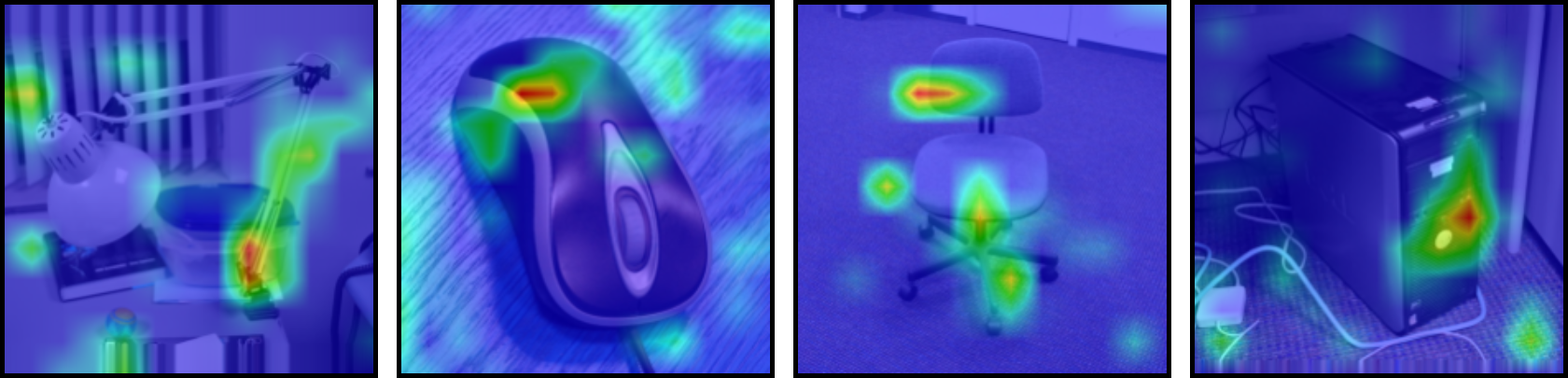}
    \caption{Visualization of certainty activation mappings using TDMDA for A$\rightarrow$D task (AlexNet). Certainty activation mappings are highlighting class-discriminative regions for target domain image similar to source domain}
  \label{fig:our_cer}
  \vspace{-1em}
\end{figure}

In figure \ref{fig:our_cer}, we can see that after using our method, certainty activation mapping now highlights class-discriminative regions in target domain. Performing only feature adaptation does not provide class-discriminative certainty activation mappings, as shown in figure~\ref{fig:dif_cer}. This depicts the effectiveness of our method in better approximating the posterior of the classifier in target domain. 

\section{Conclusion}
In this paper, we have thoroughly analyzed the problem of matching the classifier responses between the source and target domains for successful adaptation. We observe that ignoring this aspect results in higher uncertainty in the classifier and various probability distributions being quite varied between the source and target domains. As we do not have access to the target class labels, we propose a way to indirectly ensuring that the classifier distributions are matched by the triple distribution matching approach. Our thorough empirical analysis demonstrates that this does indeed result in improved performance and lower the uncertainty for domain adaptation. Moreover, as the eventual goal is classification, it is therefore more pertinent to consider the classifier response in a more comprehensive way as we show in this work.

{\small
\bibliographystyle{ieee_fullname}
\bibliography{egbib}
}

\end{document}